\documentclass[preprint,12pt,authoryear]{elsarticle}

\usepackage{amssymb}
\usepackage{amsmath}

\usepackage[whole]{bxcjkjatype} 
\usepackage{multirow}
\usepackage{subcaption}
\usepackage[table,xcdraw]{xcolor}
\usepackage{threeparttable}
\usepackage{eucal}
\usepackage{balance}
\usepackage{mathrsfs}  
\usepackage{comment}
\usepackage{algorithm}
\usepackage[noend]{algpseudocode}
\usepackage{float}
\usepackage{textcomp}
\usepackage{cancel} 
\usepackage{url}

\usepackage{xcolor}
\usepackage{ifthen}
\usepackage[normalem]{ulem}

\def\eg{\textit{e.g}\onedot} 
\def\ie{\textit{i.e}\onedot}

\newcommand{\myvec}[1]{\mathbf{#1}}

\newcommand{\myset}[1]{\mathcal{#1}}

\def\eg{{\it e.g.}}

\def\ie{{\it i.e.}}

\newboolean{showchanges}
\setboolean{showchanges}{false} 

\ifthenelse{\boolean{showchanges}}{
  \newcommand{\added}[1]{\textcolor{blue}{#1}}
  \newcommand{\deleted}[1]{\textcolor{red}{\sout{#1}}}
  \newcommand{\revised}[2]{\textcolor{red}{\sout{#1}}\textcolor{blue}{#2}}
}{
  \newcommand{\added}[1]{#1}
  \newcommand{\deleted}[1]{}
  \newcommand{\revised}[2]{#2}
}

\journal{Nuclear Physics B}

\begin{document}

\begin{frontmatter}



\title{Model Predictive Control via Probabilistic Inference:\\A Tutorial and Survey}

\author[label1,label2]{Kohei Honda}
\affiliation[label1]{organization={Nagoya University},
            addressline={Furo-cho, Chikusa-ku},
            city={Nagoya},
            postcode={464-8603},
            state={Aichi},
            country={Japan}}

\ead{honda.kohei.k3@f.mail.nagoya-u.ac.jp}
\affiliation[label2]{organization={CyberAgent AI Lab},
            addressline={Shibuya Scramble Square},
            city={Shibuya},
            postcode={150-6121},
            state={Tokyo},
            country={Japan}}



\begin{abstract}
This paper presents a tutorial and survey on Probabilistic Inference-based Model Predictive Control (PI-MPC)\deleted{for robotics}.
PI-MPC reformulates finite-horizon optimal control as inference over an optimal control distribution expressed as a Boltzmann distribution weighted by a control prior, and generates actions through variational inference.
In the tutorial part, we derive this formulation and explain action generation via variational inference, highlighting Model Predictive Path Integral (MPPI) control as a representative algorithm with a closed-form sampling update.
In the survey part, we organize existing PI-MPC research around key design dimensions, including prior design, multi-modality, constraint handling, scalability, hardware acceleration, and theoretical analysis.
This paper provides a unified conceptual perspective on PI-MPC and a practical entry point for \revised{robotics}{researchers and practitioners in robotics and other control applications}.
\end{abstract}


\end{frontmatter}

\section{Introduction}
\label{sec:introduction}

\revised{Establishing a universal control theory applicable to diverse robots and tasks has long been a fundamental goal in robotics research.}{Establishing a general control framework applicable to diverse systems and tasks has long been a fundamental goal in control and robotics research.}
Model Predictive Control (MPC) offers an intuitive and general framework \deleted{for robot control} by optimizing actions while predicting future states over a finite horizon.
To apply MPC to real-world \revised{robots and tasks}{applications}, we may need to formulate complex optimal control problems that capture the intricacies of \revised{robot}{system} dynamics and task requirements.
However, these formulations often exhibit strong nonlinearity, making them intractable for traditional numerical optimization methods widely familiar in the control field.
In many \deleted{robotic} applications \added{such as robotics}, what we need is not only strong theoretical guarantees, but also an easy-to-use optimization framework that can reliably solve a broad class of optimal control problems with minimal implementation effort.

\begin{figure}[t]
    \centering
    \includegraphics[width=1.0\linewidth]{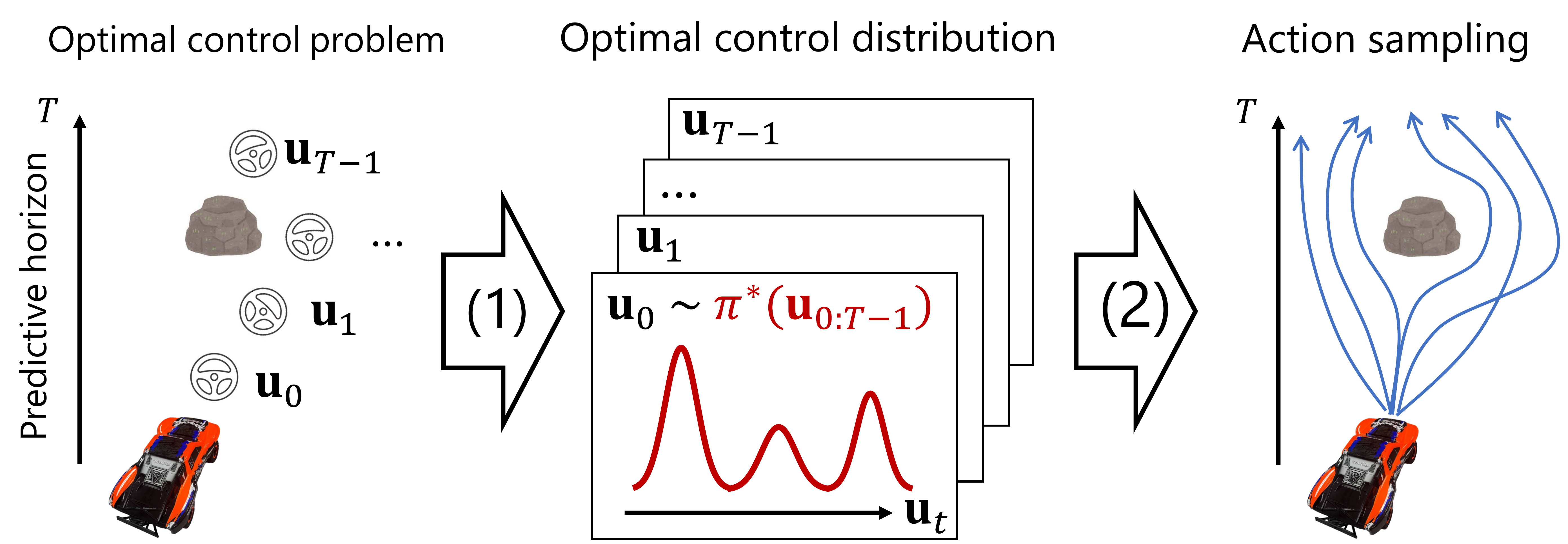}
    \caption{Overview of probabilistic inference-based model predictive control.
            The framework consists of two stages: (1) deriving the optimal control distribution from the finite-horizon optimal control problem, and (2) sampling control input sequences from this distribution for execution.
            This perspective reformulates optimal control as probabilistic inference.
            }
    \label{fig:overview}
\end{figure}

Based on the above background, \revised{probabilistic inference-based}{Probabilistic Inference-based} MPC, hereafter referred to as PI-MPC, has gained popularity for many applications recently.
As illustrated in Fig.~\ref{fig:overview}, PI-MPC represents the optimal control problem as an optimal control distribution and solves it using \revised{variational inference}{Variational Inference} (VI) techniques.
This approach belongs to the class of sampling-based MPC methods\footnote{\added{When the dynamics and cost are analytically tractable (\eg, linear-Gaussian), sampling-free inference methods can also be applied~\citep{todorov2008general, kappen2012optimal}. This tutorial focuses on sampling-based methods because they impose the fewest assumptions on cost functions and dynamics.}}, which randomly sample \revised{robot actions}{control inputs} and determine optimal behaviors from these samples.
Consequently, it is applicable to optimal control problems with arbitrary dynamics and cost functions.

We can trace attempts to solve optimal control problems through probabilistic inference back to the early 2000s, beginning with path integral control~\citep{kappen2005path} and formulating stochastic optimal control as Kullback-Leibler (KL) divergence minimization~\citep{todorov2008general, kappen2012optimal}.
It then extends to information-theoretic frameworks through Model Predictive Path Integral Control (MPPI)~\citep{williams2017information,williams2018information}, and reaches connections to Reinforcement Learning (RL) through the general framework of Control as Inference~\citep{levine2018reinforcement}.
\revised{Despite its theoretical elegance and growing impact in robotics and decision-making, its formulation is often scattered across different research communities, making it difficult for control theorists and robotic practitioners to obtain a coherent understanding.}{Despite its theoretical elegance and growing impact in robotics and broader decision-making problems, its formulation is often scattered across different research communities, making it difficult for control theorists and practitioners to obtain a coherent understanding.}
Moreover, existing tutorials tend to focus either on RL-based formulations~\citep{levine2018reinforcement} or on path integral optimal control~\citep{kazim2024recent}.
While the former is commonly framed in terms of machine learning rather than MPC, the latter is primarily developed within control theory and does not always make its connection to variational or probabilistic inference explicit.

This paper aims to provide a comprehensive tutorial and survey on PI-MPC, serving as a guide for both theorists and practitioners.
In the first half, we explain the theoretical foundations of PI-MPC, primarily based on~\citep{levine2018reinforcement,williams2018information,wang2021variational,abdolmaleki2018maximum}.
Specifically, as illustrated in Fig.~\ref{fig:overview}, we describe the approach in two steps. First, we derive the optimal control distribution from the optimal control problem and discuss its probabilistic interpretation and key characteristics.
Second, we explain how to generate actions from the optimal control distribution using VI techniques, taking MPPI as a representative example\footnote{We open-source a PyTorch-based MPPI implementation, available at \url{https://github.com/kohonda/mppi_playground}}.
In the latter half, we survey key challenges, representative methods, and recent advances in PI-MPC, including fundamental topics and cutting-edge research directions.
\section{Tutorial on Probabilistic Inference-Based Model Predictive Control}
\label{sec:tutorial}

\subsection{Definition of Optimal Control Problem}
\revised{Our objective is to find the optimal control input (action) sequence $\myvec{u}^*_{0:T-1}$ by solving the following optimal control problem over a finite time horizon $T$:}
{Our objective is to find the optimal control input sequence $\myvec{u}^*_{0:T-1}$ and its corresponding state trajectory $\myvec{x}^*_{0:T}$ over a finite horizon $T$ by solving the following optimal control problem:}
\begin{subequations}
\begin{align}
\min_{\added{\myvec{x}_{0:T},}\myvec{u}_{0:T-1}}\; &J\left( \myvec{x}_{0:T}, \myvec{u}_{0:T-1} \right), \label{eq:cost_func}\\
\text{s.t.}\; &\myvec{x}_{t+1}= \myvec{f} (\myvec{x}_t, \myvec{u}_t), \quad ^\forall t \in  \{ 0,\dots,T-1 \}.  \label{eq:dynamics}
\end{align}
\label{eq:ocp}
\end{subequations}
Here, $J$ is the cost function; $\myvec{x}_t$ and $\myvec{u}_t$ denote the state and control input at time step $t$, respectively.
We write the control sequence as $\myvec{u}_{0:T-1} = [\myvec{u}_0, \dots, \myvec{u}_{T-1}]$ and the corresponding state sequence as $\myvec{x}_{0:T} = [\myvec{x}_0, \dots, \myvec{x}_T]$.
The dynamics are given by $\myvec{f}$, and $\myvec{x}_0$ is the observed initial state.
\added{Although Eq.~(\ref{eq:ocp}) is formally written as an optimization over both $\myvec{x}_{0:T}$ and $\myvec{u}_{0:T-1}$, the state trajectory is determined by the control sequence through forward rollout of the dynamics $\myvec{f}$ once $\myvec{x}_0$ is given. 
Therefore, the PI-MPC framework focuses on optimizing the control sequence $\myvec{u}_{0:T-1}$, with the state trajectory $\myvec{x}_{0:T}$ implicitly defined by the dynamics.}
\added{In the MPC setting, Eq.~(\ref{eq:ocp}) is solved repeatedly in a receding-horizon manner. At each time step, the current state is observed, an optimal control sequence is computed over the predictive horizon, only the first control input is applied, and the optimization is solved again at the next time step after shifting the horizon forward.}

\added{The cost function $J$ is left general: it may include running costs, terminal costs, or penalty terms for constraint violation. In classical MPC, a carefully designed terminal cost and terminal constraint set are often required to guarantee closed-loop stability~\citep{rawlings2017mpc}. PI-MPC does not impose such structural requirements on $J$, which broadens its applicability but means that formal stability guarantees are not provided by the framework itself; see Section~\ref{sec:survey_theory} for available theoretical results.}

\deleted{Given $J$ and $\myvec{f}$, the controller solves Eq.~(\ref{eq:ocp}) online and applies the resulting action sequence $\myvec{u}^*_{0:T-1}$. When $J$ and $\myvec{f}$ admit simple structures (\eg, quadratic objectives and differentiable dynamics), numerical optimization can be efficient.}
\added{On the other hand, this generality can make numerical optimization of Eq.~(\ref{eq:ocp}) challenging within the classical MPC framework. When $J$ and/or $\myvec{f}$ have simple structures (\eg, quadratic costs and smooth differentiable dynamics), the optimization can often be solved efficiently. However, when $J$ and/or $f$ are highly nonlinear, nonconvex, or nondifferentiable, or when stochastic dynamics are considered, real-time optimization becomes more challenging. In such cases, classical gradient- or Hessian-based MPC solvers may suffer from sensitivity to initialization, poor numerical conditioning, and convergence to local minima rather than the global optimum. If $J$ or $f$ is nondifferentiable, these solvers may not be directly applicable at all, unless additional smoothing or subgradient-based approximations are introduced. Stochastic dynamics further complicate the problem by making trajectory evaluations noisy and increasing the computational burden.}

\subsection{Motivations for Probabilistic Inference-Based MPC}

When numerical optimization proves infeasible, our remaining option is computationally intensive sampling-based methods.
The simplest approach is the random shooting method: we sample many candidate control sequences $\myvec{u}_{0:T-1}$, roll out the resulting trajectories using Eq.~(\ref{eq:dynamics}), evaluate each trajectory cost via Eq.~(\ref{eq:cost_func}), and choose the sequence with the lowest cost.
\added{In its simplest form, random shooting performs a single round of sampling and selection without iterative refinement}\footnote{\added{Variants, such as the Cross-Entropy Method (CEM)~\citep{botev2013cross}, improve upon this by iteratively resampling from an updated distribution fitted to the top-performing candidates, but such iterative update procedure increases computational cost.}}.

\begin{figure}[t]
    \centering
    \begin{minipage}{0.48\linewidth}
        \centering
        \includegraphics[width=0.8\linewidth]{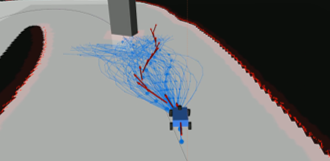}
        \subcaption{Random shooting}
    \end{minipage}
    \hfill
    \begin{minipage}{0.48\linewidth}
        \centering
        \includegraphics[width=0.8\linewidth]{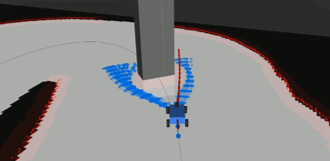}
        \subcaption{MPPI}
    \end{minipage}
    \caption{Comparison of sampling-based MPC methods in a vehicle obstacle avoidance task~\citep{honda2024stein}. 
            Random shooting selects a single best trajectory from samples, resulting in poor sample efficiency, whereas MPPI estimates an optimal control distribution and achieves significantly improved efficiency through distribution-level inference.
            }
    \label{fig:random_shooting_vs_mppi}
\end{figure}

As shown in Fig.~\ref{fig:random_shooting_vs_mppi}, random shooting is highly sample-inefficient and quickly suffers from the curse of dimensionality: the number of required samples grows exponentially with the input dimension.
\added{With a fixed computational budget, this means that the probability of finding a near-optimal solution degrades rapidly as dimensionality increases, rather than the problem becoming computationally intractable in a strict sense.}

To overcome the sample efficiency challenges, we therefore introduce a probabilistic inference framework. Specifically, as illustrated in Fig.~\ref{fig:overview}, we approximate the probabilistic distribution of optimal control inputs (hereafter referred to as the \emph{optimal control distribution}) using samples to generate control input sequences from this distribution.
Rather than searching for a single best sequence among random samples, we infer a distribution over good control sequences; this \emph{distribution-level} view can substantially improve sample efficiency (Fig.~\ref{fig:random_shooting_vs_mppi}).
Beyond sample efficiency, casting control as distribution optimization offers additional benefits:
\begin{itemize}
  \item Most computations are embarrassingly parallel, making CPU/GPU parallelization straightforward.
  \item \deleted{It naturally represents stochastic variability in behavior, which is beneficial for exploration and enables effective data augmentation, as described in Section~\ref{sec:survey_connections}.} \added{It naturally represents stochastic variability in behavior. This variability is beneficial for exploration in model-based reinforcement learning and for effective data augmentation, as described in Section~\ref{sec:survey_connections}. For example, in model-based reinforcement learning, diverse trajectories can improve the training efficiency~\citep{okada2020variational}. In pure control settings, the distribution is often collapsed to its mean for execution (\eg, Eq. (\ref{eq:mppi_mu})), but maintaining distributional diversity during the planning phase helps avoid premature convergence to local optima~\citep{lambert2021stein}.}
  \item \revised{The overall pipeline can be implemented end-to-end in a differentiable manner (\eg, in PyTorch), enabling seamless integration with learning-based models and gradient-based optimization of both control and model parameters.}{When the dynamics model and cost function are implemented as differentiable functions (\eg, neural networks), the entire computation pipeline, as described in Section~\ref{sec:mppi_derivation}, can be executed within an automatic differentiation framework such as PyTorch. This enables gradient-based optimization of model parameters or cost function parameters through the PI-MPC pipeline, facilitating joint learning and control as discussed in Section~\ref{sec:survey_connections}.}
\end{itemize}

\subsection{Framework of Probabilistic Inference-Based MPC}

In this section, we outline the basic PI-MPC framework.
As shown in Fig.~\ref{fig:overview}, the scheme involves two key steps: deriving the optimal control distribution from the optimal control problem in Eq.~(\ref{eq:ocp}) and then generating control inputs to apply to the \revised{robot}{system} from this optimal control distribution.
Thus, two fundamental questions arise:
\begin{enumerate}
  \item How is the optimal control distribution represented, and what are its characteristics?
  \item How do we generate control inputs to apply to the \revised{robot}{system} from the optimal control distribution?
\end{enumerate}
The following exposition focuses on these two fundamental questions.

\subsubsection{Derivation of the Optimal Control Distribution}

We begin by deriving the optimal control distribution from the optimal control problem in Eq.~(\ref{eq:ocp}).
We first state the key result: under mild assumptions\footnote{\added{The derivation relies on the Markov structure of the dynamics in Eq.~(\ref{eq:dynamics}) and assumes that the optimality likelihood takes the Boltzmann form in Eq.~(\ref{eq:boltzmann_distribution}).}}, the probability density function of the optimal control distribution $\pi^*$ takes the following simple form:
\begin{align}
& \underbrace{\pi^*(\myvec{u}_{0:T-1})}_{\text{\added{Optimal control distribution}}} = Z^{-1} \underbrace{\exp\left(-\lambda^{-1} J_{\tau}(\myvec{u}_{0:T-1})\right)}_{\text{Boltzmann distribution}} \underbrace{p(\myvec{u}_{0:T-1})}_{\text{Prior distribution}}. \label{eq:optimal_control_distribution}
\end{align}
Here, $Z$ is the normalization constant, $\lambda$ is the temperature parameter, $J_{\tau}(\myvec{u}_{0:T-1})$ is the trajectory cost obtained from the control input $\myvec{u}_{0:T-1}$ via Eq.~(\ref{eq:ocp}), and $p(\myvec{u}_{0:T-1})$ represents the prior distribution over control inputs.
Thus, the optimal control distribution is proportional to the product of a Boltzmann distribution and a prior distribution\footnote{A similar expression appears in Direct Preference Optimization~\citep{rafailov2023direct}, which has been used for aligning large language models.}. The Boltzmann term assigns higher probability to lower-cost trajectories; we will discuss it in more detail later.

\added{The prior distribution $p(\myvec{u}_{0:T-1})$ represents the default behavior of the controller. 
Since the optimal control distribution $\pi^*$ is proportional to the product of the Boltzmann distribution and the prior distribution, $\pi^*$ assigns high probability to control sequences that are both low-cost and probable under the prior. 
In other words, the prior determines the region over which cost-driven optimization is effective, and we can shape it to bias the search toward promising regions of the control space.
Its design thus significantly affects the quality of the resulting control, as discussed in Section~\ref{sec:prior_distribution} and Section~\ref{sec:survey_prior_distributions}.}

\revised{We now derive this expression in three steps.}{We now derive Eq.~(\ref{eq:optimal_control_distribution}) in three steps.}
\paragraph{Step 1: Graphical Model Representation of the Optimal Control Problem}  

\begin{figure}[t]
    \centering
    \includegraphics[width=1.0\linewidth]{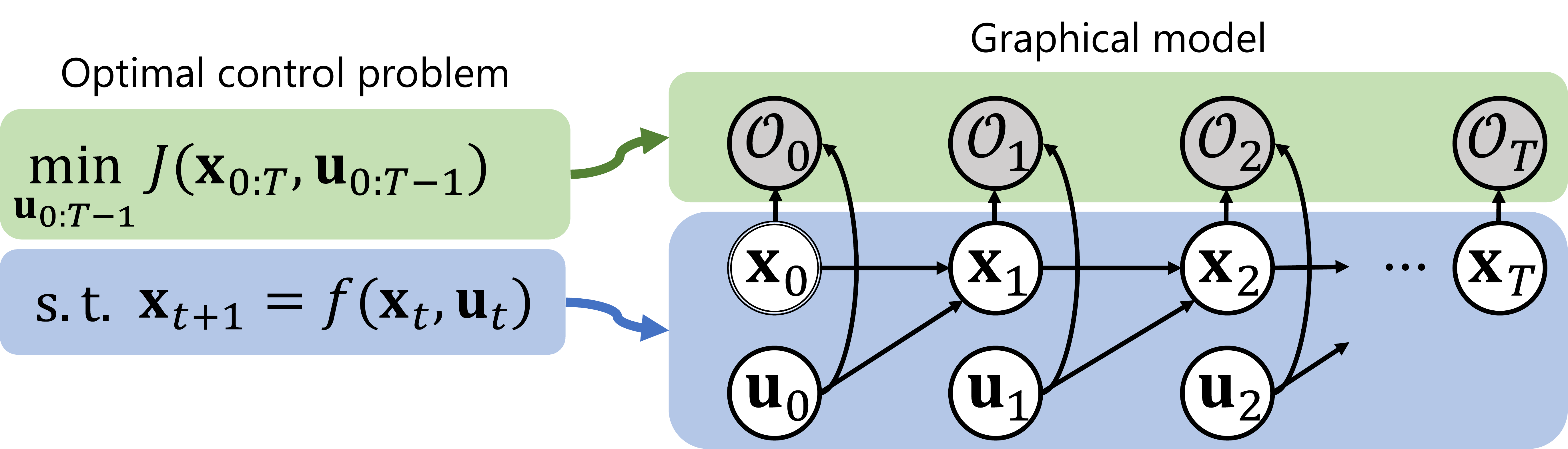}
    \caption{Graphical-model representation of the finite-horizon optimal control problem.
              State transitions depend on current states and control inputs, while optimality variables encode trajectory cost as probabilistic likelihood.
              This formulation enables variational inference of the optimal control distribution.}
    \label{fig:graphical_model}
\end{figure}

We rewrite the optimal control problem in Eq.~(\ref{eq:ocp}) as a graphical model.
In a graphical model, nodes denote random variables and edges encode conditional dependencies.
Figure~\ref{fig:graphical_model} shows the resulting factorization.
In particular, the next state $\myvec{x}_{t+1}$ depends on the current state $\myvec{x}_t$ and control $\myvec{u}_t$, which corresponds to the dynamics in Eq.~(\ref{eq:dynamics}).

The key additional node is $\myset{O}_t$, the \emph{optimality variable}.
It is a virtual binary variable indicating whether the state--action pair at time $t$ is ``optimal'' ($\myset{O}_t=1$) or not ($\myset{O}_t=0$), and it will be used to encode the cost function in Eq.~(\ref{eq:cost_func}) as a likelihood term.

Based on this model, we define the trajectory $\tau = [\myvec{x}_{0:T}, \myvec{u}_{0:T-1}]$ and the sequence of optimality variables $\myset{O}_{0:T} = [\myset{O}_0, \dots, \myset{O}_T]$.
Conditioning on the event that all time steps are optimal, Bayes' rule yields the posterior trajectory distribution $p(\tau \mid \myset{O}_{0:T}=\myvec{1})$:
\begin{align}
p(\tau \mid \myset{O}_{0:T} = \myvec{1}) 
& \propto  \underbrace{p(\myset{O}_{0:T} = \myvec{1} \mid \tau)}_{\text{Optimality likelihood}} \times \underbrace{p(\tau)}_{\text{Trajectory distribution}}  \nonumber \\
& = p(\myset{O}_{0:T}=\myvec{1} \mid \tau) p(\myvec{x}_0) \prod_{t=0}^{T-1}  p(\myvec{x}_{t+1} \mid \myvec{x}_t, \myvec{u}_t) p(\myvec{u}_t), \label{eq:optimal_trajectory_distribution}
\end{align}
where $p(\myset{O}_{0:T} = \myvec{1} \mid \tau)$ is the \emph{optimality likelihood}, which assigns higher probability to trajectories that are more ``optimal'' under the cost.

\begin{figure}[t]
    \centering
    \includegraphics[width=1.0\linewidth]{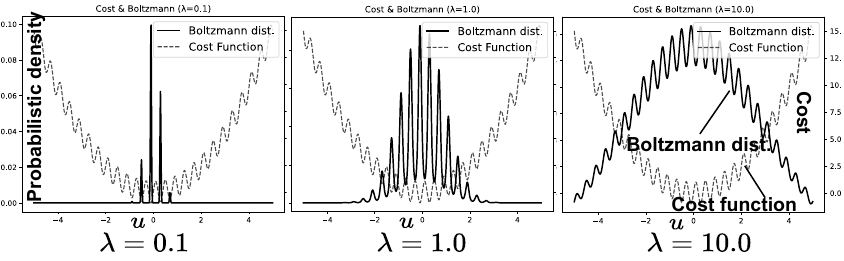}
    \caption{Relationship between the cost function and the Boltzmann distribution.
            The Boltzmann distribution assigns higher probability to trajectories with lower costs, and its sharpness is governed by the temperature parameter $\lambda$.
            Smaller $\lambda$ concentrates probability around the minimum-cost region, whereas larger $\lambda$ yields a smoother distribution.
            The example cost function is $J(u)=0.6u^2 + \sin(5\pi u)$ with time horizon $T=1$.}
    \label{fig:cost_and_boltzman}
\end{figure}

\paragraph{Step 2: Approximating the Optimal Control Distribution via Kullback-Leibler Divergence Minimization}
We modeled the optimal trajectory distribution $p(\tau \mid \myset{O}_{0:T} = \myvec{1})$ in Eq.~(\ref{eq:optimal_trajectory_distribution}).
However, since the virtual binary variable $\myset{O}_{0:T}$ cannot be computed in advance, this distribution is intractable.
Additionally, since the prior distribution $p(\myvec{u}_{0:T-1})$ included in $p(\tau)$ may contain system-side input noise, it is not necessarily directly controllable by us.
We therefore introduce a controllable trajectory distribution $\pi(\tau)$ and use variational inference (VI) to make it close to $p(\tau \mid \myset{O}_{0:T} = \myvec{1})$.
Specifically, we minimize the KL divergence $\mathbb{D}_{\rm{KL}}$:
\begin{align}
  \pi^*(\tau) 
  & = \min_{\pi} \mathbb{D}_{\rm{KL}}\left(\pi(\tau) \parallel p(\tau \mid \myset{O}_{0:T} = \myvec{1}) \right) \nonumber \\
  & = \min_{\pi} \mathbb{E}_{\pi(\tau)}\left[ \log \pi(\tau) - \log p(\tau \mid \myset{O}_{0:T} = \myvec{1}) \right] \nonumber \\
  & = \min_{\pi}  \{ \mathbb{E}_{\pi(\tau)}[-\log p(\myset{O}_{0:T} = \myvec{1} \mid \tau)]  + \mathbb{E}_{\pi(\tau)}\left[ \log \pi(\tau) - \log p(\tau) \right]  \}, \label{eq:kl_divergence_min_wip}
\end{align}
where the final transformation uses Eq.~(\ref{eq:optimal_trajectory_distribution}).
If we assume a Boltzmann form for the optimality likelihood $p(\myset{O}_{0:T} = \myvec{1} \mid \tau)$, then marginalizing out the states yields:
\begin{align}
\pi^*(\myvec{u}_{0:T-1})
& = \min_{\pi} \{ \lambda^{-1} \overbrace{\mathbb{E}_{\pi(\myvec{u}_{0:T-1})}\left[  J_{\tau}(\myvec{u}_{0:T-1}) \right]}^{\text{Expected cost function}} \nonumber \\
& \;\; \;\;  + \underbrace{\mathbb{D}_{\rm{KL}}\left(\pi(\myvec{u}_{0:T-1}) \parallel p(\myvec{u}_{0:T-1}) \right)}_{\text{Regularization term for deviation from prior dist.}} \},
\label{eq:kl_divergence_min}
\end{align}
where, with normalization constant $\eta$:
\begin{align}
  & p(\myset{O}_{0:T} = \myvec{1} \mid \tau) = \eta^{-1} \exp\left(-\lambda^{-1} J_{\tau}(\myvec{u}_{0:T-1})\right), \label{eq:boltzmann_distribution} \\
  & J_{\tau}(\myvec{u}_{0:T-1}) = \mathbb{E}_{p(\myvec{x}_{0:T} \mid \myvec{u}_{0:T-1})}\left[ J(\myvec{x}_{0:T}, \myvec{u}_{0:T-1}) \right], \label{eq:traj_cost}
\end{align}
are substituted into Eq.~(\ref{eq:kl_divergence_min_wip})\footnote{Equation~(\ref{eq:kl_divergence_min}) can also be derived via evidence lower bound minimization.}.
Equation~(\ref{eq:boltzmann_distribution}) defines the Boltzmann distribution, which assigns higher probability to lower-cost trajectories.
As illustrated in Fig.~\ref{fig:cost_and_boltzman}, the temperature parameter $\lambda$ controls its sharpness: a smaller $\lambda$ concentrates mass near the minimum-cost region.
Although the Boltzmann form can be motivated by maximum-entropy arguments, it is not the only choice; alternative likelihood models are possible\footnote{For example, the CEM uses a thresholded likelihood~\citep{botev2013cross}.}.
We use the Boltzmann form because it leads to a simple and practical inference procedure.

Equation~(\ref{eq:kl_divergence_min}) shows that $\pi^*$ trades off two terms: the expected cost and the KL divergence to the prior.
The temperature $\lambda$ determines the relative weight of these terms.
In other words, $\pi^*(\myvec{u}_{0:T-1})$ favors low expected cost while remaining close to the prior distribution.

\begin{figure}[t]
    \centering
    \includegraphics[width=1.0\linewidth]{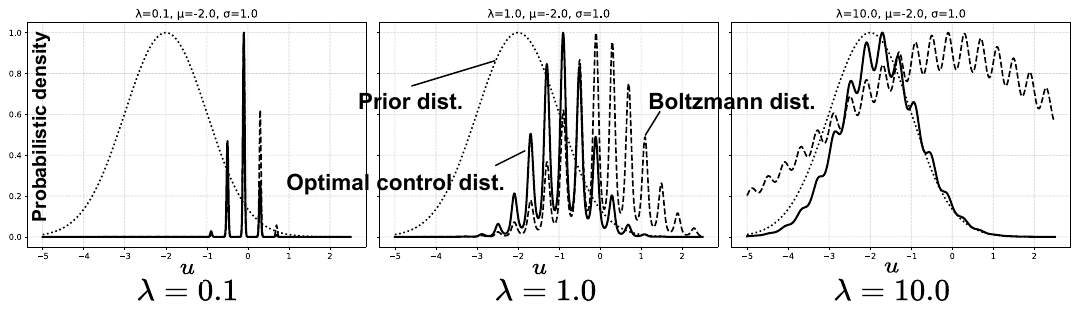}
    \caption{Temperature $\lambda$-dependent variation of the optimal control distribution.
            The optimal control distribution is formed by the product of the Boltzmann distribution and the prior distribution.
            Larger temperature values increase the influence of the prior, yielding smoother distributions closer to the prior, whereas smaller temperatures emphasize cost optimality and produce sharper modes.
            The cost function is the same as in Fig.~\ref{fig:cost_and_boltzman}, and the prior distribution is $p(u)=\mathcal{N}(-2.0,1.0)$.
            }
    \label{fig:ocp_lambda}
\end{figure}

\paragraph{Step 3: Computing the Optimal Control Distribution via the Method of Lagrange Multipliers}

Finally, we derive Eq.~(\ref{eq:optimal_control_distribution}) from the KL-minimization problem in Eq.~(\ref{eq:kl_divergence_min}) using a Lagrange multiplier $\alpha$.
We minimize Eq.~(\ref{eq:kl_divergence_min}) subject to the normalization constraint $\int \pi(\myvec{u}_{0:T-1}) d\myvec{u}_{0:T-1} = 1$.
Define the Lagrangian
\begin{align}
& \mathcal{L}(\pi, \alpha) \nonumber \\
& = \lambda^{-1} \mathbb{E}_{\pi(\myvec{u}_{0:T-1})}\left[  J_{\tau}(\myvec{u}_{0:T-1}) \right] + \mathbb{D}_{\rm{KL}}\left(\pi(\myvec{u}_{0:T-1}) \parallel p(\myvec{u}_{0:T-1}) \right) + \alpha \left( \int \pi d\myvec{u}_{0:T-1} - 1 \right) \nonumber \\
\end{align}
and set its functional derivative with respect to $\pi$ to zero:
\begin{align}
& \frac{\partial \mathcal{L}}{\partial \pi}
= \lambda^{-1}J_{\tau}+ \log \frac{\pi^*}{p} -1 + \alpha = 0 \nonumber \\
\Leftrightarrow \;\; & \pi^* = \exp (1-\alpha) \exp\left(-\lambda^{-1} J_{\tau} \right) p(\myvec{u}_{0:T-1}). \nonumber
\end{align}
Enforcing the normalization constraint gives
$$\int \pi^* d\myvec{u}_{0:T-1} = \exp(1-\alpha) \int \exp \left(-\lambda^{-1} J_{\tau} \right) p(\myvec{u}_{0:T-1}) d\myvec{u}_{0:T-1} = 1,$$ 
which implies
$$
\exp(1-\alpha) = \left[\int \exp\left(-\lambda^{-1} J_{\tau} \right) p(\myvec{u}_{0:T-1}) d\myvec{u}_{0:T-1}\right]^{-1} =  Z^{-1}.
$$
Therefore, we recover Eq.~(\ref{eq:optimal_control_distribution}).

\subsection{Analysis of the Optimal Control Distribution $\pi^*$}\label{sec:discussion}
Having derived Eq.~(\ref{eq:optimal_control_distribution}), we now briefly discuss its interpretation and key properties.

\subsubsection{{Influence of the Prior Distribution on the Optimal Control Distribution}}\label{sec:prior_distribution}

As shown in Eq.~(\ref{eq:optimal_control_distribution}), the optimal control distribution is expressed as the product of the Boltzmann distribution and the control input prior distribution.
Thus, ``optimal actions'' are not simply those that minimize the cost in Eq.~(\ref{eq:cost_func}); they also reflect the prior distribution (Fig.~\ref{fig:ocp_lambda}).

\added{From a control-theoretic viewpoint, the prior distribution plays a role analogous to the initial guess in iterative optimization. 
In the receding-horizon setting, reusing the posterior from the previous time step as the current prior distribution provides a natural warm-start (Section~\ref{sec:mppi_derivation}). 
More broadly, domain knowledge such as actuator limits or pre-trained policies can be encoded in the prior distribution to improve sample efficiency, as discussed in Section~\ref{sec:survey_prior_distributions}.}

\citet{kappen2005path} illustrates this characteristic with the analogy of a ``drunken spider crossing a pond.''
Imagine a spider attempting to cross a pond via a narrow bridge.
When sober (\ie, when the prior has small variance), the best strategy is to take the shortest route across the bridge.
When drunk (\ie, when the prior variance is large), crossing the narrow bridge becomes risky, and a longer detour around the pond may be preferable.
The lesson is that optimal actions depend not only on the cost function but also on the prior distribution.

Note that models that use Boltzmann distributions as optimal distributions for sample generation are called Energy-Based Models (EBM), well studied in machine learning~\citep{lecun2006tutorial}.
Compared with standard EBMs, Eq.~(\ref{eq:optimal_control_distribution}) includes an explicit action prior, which is induced by the system dynamics in Eq.~(\ref{eq:dynamics}).

\subsubsection{{Decision-Making Through Symmetry Breaking}}\label{sec:symmetry_breaking}

\begin{figure}[t]
    \centering
    \includegraphics[width=1.0\linewidth]{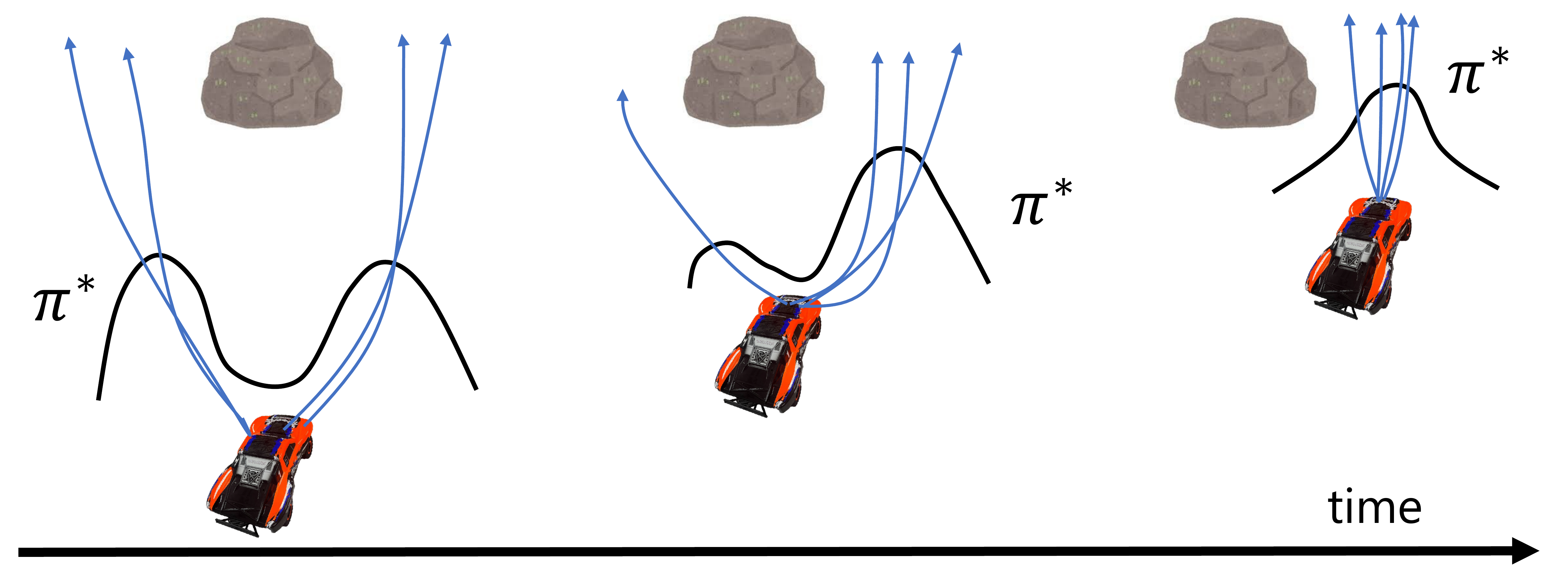}
    \caption{Decision-making through symmetry breaking in the optimal control distribution.
             When multiple avoidance paths exist, the optimal control distribution initially exhibits multiple modes with nearly equal probability.
             As time progresses and the prior variance effectively decreases, stochastic bias toward one direction causes the distribution to collapse into a single mode, resulting in delayed but decisive action selection.
            }
    \label{fig:symmetry_breaking}
\end{figure}

One intriguing phenomenon arising from the prior distribution is ``symmetry breaking''~\citep{kappen2005path}.
Symmetry breaking refers to the phenomenon where the number of modes in the optimal control distribution decreases over time as the noise level of the prior distribution diminishes.

Consider the obstacle-avoidance example in Fig.~\ref{fig:symmetry_breaking}.
When the robot is far from the obstacle, the optimal control distribution is typically bimodal, corresponding to left and right avoidance paths.
The two modes often have nearly equal probability, so the controller does not commit strongly to either direction.
As the robot approaches the obstacle, random fluctuations in the prior can bias the distribution toward one side, and the optimal control distribution (constrained by the prior) eventually collapses to a single mode.

From a control perspective, this mode convergence through symmetry breaking represents delayed decision-making.
In deterministic control methods, ``early decisions'' to proceed left or right occur even far from obstacles. In contrast, probabilistic inference-based methods tend to postpone decisions until reasonably close to obstacles.

\subsubsection{{Relationship Between Temperature Parameter and Optimal Control Distribution}}\label{sec:temperature_parameter}

As shown in Eq.~(\ref{eq:optimal_control_distribution}), the optimal control distribution depends on the temperature parameter $\lambda$.
The temperature parameter originates from Eq.~(\ref{eq:boltzmann_distribution}) and, as shown in Eq.~(\ref{eq:kl_divergence_min}), controls the weight between the expected cost and the regularization term for deviation from the prior.

Figure~\ref{fig:ocp_lambda} illustrates the effect of $\lambda$.
Larger values of $\lambda$ increase the influence of the prior, yielding smoother distributions closer to the prior.
Smaller values emphasize the Boltzmann term, producing sharper distributions concentrated near low-cost regions.
As extreme cases, $\lambda \to 0$ yields deterministic optimal actions that minimize the cost function, while $\lambda \to \infty$ yields stochastic optimal actions following only the prior distribution.
Thus, the temperature parameter tunes the balance between deterministic and stochastic characteristics of the optimal control distribution.

The temperature parameter is a hyperparameter that must be tuned for each application.
Smaller $\lambda$ encourages more aggressive cost-seeking behavior, which can improve performance but may reduce sample efficiency and robustness to disturbances.
Larger $\lambda$ places more weight on the prior, which can improve sample efficiency and robustness but may become overly conservative.
In many works, a fixed temperature parameter is manually tuned, though adaptive adjustment methods have also been proposed~\citep{watson2023inferring,pezzato2025sampling}.

\subsubsection{{Factors Governing Sample Complexity}} \label{sec:sample_complexity}

As we discuss in Section~\ref{sec:mppi_derivation}, in practice, we approximate the optimal control distribution in Eq.~(\ref{eq:optimal_control_distribution}) using Monte Carlo sampling.
Accordingly, it is desirable for the target distribution to be well approximated with as few samples as possible.
The number of samples needed to obtain a good approximation depends strongly on the distribution's shape; this is often referred to as \emph{sample complexity}.

In general, sample complexity increases as the distribution becomes more sharply peaked, requiring more samples\footnote{Intuitively, it is easier to hit a broad ``winning region'' with random samples than to pinpoint a single ``correct answer.''}.
\added{For example, in the extreme case of $\lambda \to 0$, the optimal control distribution concentrates at the minimum-cost point, which excessively increases sample complexity. Even in simple problems such as linear-quadratic control, this would make it difficult for sampling-based methods to find the optimal solution}\footnote{\added{In analytically tractable cases such as linear-quadratic control, the same probabilistic viewpoint can still be formulated. However, when closed-form or more direct solution methods are available, approximating the optimal control distribution by sampling is usually computationally inefficient compared with those alternatives.}}.

According to Eq.~(\ref{eq:optimal_control_distribution}), the main factors determining the optimal control distribution's shape are:
\begin{itemize}
  \item Temperature parameter $\lambda$: as discussed in Section~\ref{sec:temperature_parameter}, smaller $\lambda$ yields sharper distributions and thus higher sample complexity.
  \item Prior distribution $p(\myvec{u}_{0:T-1})$: tighter priors (\ie, less exploration noise) also increase sample complexity.
  \item Cost landscape $J_\tau$: when the cost changes steeply (\eg, many samples collide with obstacles), the effective distribution becomes harder to approximate, increasing sample complexity.
\end{itemize}
\deleted{Because these factors interact, it is difficult to predict the required number of samples a priori; in practice, using more samples is often beneficial when computationally feasible.}
\added{Because these factors interact, it is difficult to predict the required number of samples a priori.
Some works have theoretically analyzed the required number of samples to bound the expected estimation error of MPPI as a function of the cost function and the covariance of the prior distribution~\citep{Yoon2022-jb,Tao2022-dj}.
However, these analyses do not consider the influence of the temperature parameter and tend to recommend conservative sample numbers since they provide only probabilistic bounds.
Therefore, although partial theoretical bounds are available, estimating the required number of samples a priori in a general setting remains an open problem.
In practice, it is often beneficial to use more samples when computationally feasible or to monitor the effective sample size~\citep{kong1992note} and dynamically adjust the number of samples~\citep{watson2023inferring}. }

\subsection{Control Input Generation from the Optimal Control Distribution: The MPPI Example} \label{sec:mppi_derivation}

The preceding sections focused on how the optimal control distribution is represented and its characteristics.
As shown in Fig.~\ref{fig:overview}, we now focus on how to generate control inputs from the optimal control distribution---the essential step for actual control.

Despite having a mathematical expression for the optimal control distribution in Eq.~(\ref{eq:optimal_control_distribution}), directly sampling from it computationally is challenging.
This is because the distribution can be highly complex depending on the cost function and prior, or may be ill-defined until a specific control input sequence is determined.
We therefore again employ a VI approach.
Specifically, we approximate the optimal control distribution $\pi^*$ using a parameterized variational distribution $\pi_{\theta}$:
\begin{align}
  & \myvec{u}_{0:T-1}^* \sim \pi_{\theta}^* = \min_{\theta} \mathbb{D}_{\rm{KL}}\left(  \pi^* \parallel \pi_{\theta} \right), \label{eq:variational_inference}
\end{align}
and seek the variational distribution $\pi_{\theta}$ that achieves this minimum. The specific choice of $\pi_{\theta}$ varies by method, as discussed in Section~\ref{sec:multimodal_vi_dist}.

The most widely used PI-MPC method is MPPI~\citep{williams2018information}.
MPPI employs a multivariate Gaussian distribution with fixed covariance $\Sigma$ as the variational distribution.
Specifically, setting $\theta = \mu_{0:T-1}=[\mu_0, \dots, \mu_{T-1}]$ and 
$$
\pi_{\mu_{0:T-1}} = \prod_{t=0}^{T-1} \mathcal{N}(\mu_t, \Sigma), 
$$
where
\begin{align}
\mu_{0:T-1}
& = \min_{\mu_{0:T-1}} \mathbb{D}_{\rm{KL}}\left(  \pi^*\parallel \prod_{t=0}^{T-1} \mathcal{N}(\mu_t, \Sigma) \right) \nonumber \\
& =  \min_{\mu_{0:T-1}} \mathbb{E}_{\pi^*}\left[ \cancel{\log \pi^*} - \log \prod_{t=0}^{T-1} \mathcal{N}(\mu_t, \Sigma) \right] \nonumber \\
& = \max_{\mu_{0:T-1}} \mathbb{E}_{\pi^*} \underbrace{\left[\sum_{t=0}^{T-1} \log \mathcal{N}(\mu_t, \Sigma)\right]}_{\text{Concave}}  \label{eq:convex_objective} \\
& = \max_{\mu_{0:T-1}} \int \pi^* \log \mathcal{N}(\mu_t, \Sigma) d\myvec{u}_{0:T-1}. \label{eq:mppi_objective}
\end{align}
Since Eq.~(\ref{eq:convex_objective}) is concave, the global optimal solution $\mu_{0:T-1}^*$ can be obtained via the stationarity condition.
Setting $\mathcal{L} = \int \pi^* \log \mathcal{N}(\mu_t, \Sigma) d\myvec{u}_{0:T-1}$:
\begin{align}
\frac{\partial \mathcal{L}}{\partial \mu_t} & = \int \pi^* (\myvec{u}_{0:T-1} - \mu_{0:T-1}^*) d\myvec{u}_{0:T-1} = 0 \nonumber \\
\Leftrightarrow \;\;
\mu_{0:T-1}^*
&= \frac{\int \myvec{u}_{0:T-1} \exp(-\lambda^{-1} J_{\tau}) p(\myvec{u}_{0:T-1}) d\myvec{u}_{0:T-1}}{\int \exp(-\lambda^{-1} J_{\tau}) p(\myvec{u}_{0:T-1}) d\myvec{u}_{0:T-1}} \nonumber \\
\phantom{\Leftrightarrow \mu_{0:T-1}^*} \;
& = \frac{\mathbb{E}_{p(\myvec{u}_{0:T-1})}\left[ \myvec{u}_{0:T-1} \exp(-\lambda^{-1} J_{\tau}(\myvec{u}_{0:T-1})) \right]}{\mathbb{E}_{p(\myvec{u}_{0:T-1})}\left[ \exp(-\lambda^{-1} J_{\tau}(\myvec{u}_{0:T-1})) \right]}, \label{eq:mppi_mu}
\end{align}
where we use Eq.~(\ref{eq:optimal_control_distribution}).
The expectation $\mathbb{E}_{p(\myvec{u}_{0:T-1})}[\cdot]$ in Eq.~(\ref{eq:mppi_mu}) can be computed numerically via Monte Carlo methods, with the prior distribution $p(\myvec{u}_{0:T-1})$ for random sampling typically taken as the optimal solution from the previous time step, \ie, 
$$p(\myvec{u}_{0:T-1}) = \mathcal{N}(\mu_{0:T-1}^{\rm{prev}}, \Sigma).$$

Finally, the MPPI algorithm consists of just three remarkably simple steps:
\begin{enumerate}
  \item Sample $K$ control input sequences from the prior distribution: $$\myvec{u}_{0:T-1}^{k} \sim \mathcal{N}(\mu_{0:T-1}^{\rm{prev}}, \Sigma)\;\; (k=\{1, \dots, K\})$$
  \item Forward rollout and trajectory cost evaluation using Eq.~(\ref{eq:traj_cost}): $$J_{\tau}^{k} = J_{\tau}(\myvec{u}_{0:T-1}^{k}) \;\; (k=\{1, \dots, K\})$$
  \item Update the mean via the Monte Carlo approximation of Eq.~(\ref{eq:mppi_mu}):
  \begin{align}
    \mu_{0:T-1}^* = \sum_{k=1}^{K} \left \{ \rm{Softmax}(-\lambda^{-1}J_{\tau}^k) \myvec{u}_{0:T-1}^k\right \} \label{eq:mppi_algorithm}
  \end{align}
\end{enumerate}
Note that Eq.~(\ref{eq:mppi_mu}) can be implemented using the Softmax weighting in Eq.~(\ref{eq:mppi_algorithm}), \ie, a weighted average of the sampled control sequences.
Moreover, steps (1) and (2) are independent operations for each sample $k$, enabling straightforward GPU parallelization via PyTorch and similar frameworks.
If input constraints must be enforced, we can clamp each sampled control sequence between steps (1) and (2); this clamping can be treated as part of the dynamics.

\added{
The computational cost of MPPI scales as $\myset{O}(K T c_f)$ where $K$ is the number of samples, $T$ is the horizon length, and $c_f$ is the cost of a single forward dynamics evaluation. 
Since the $K$ rollouts are independent, they parallelize trivially across CPU/GPU threads; the effective wall-clock time thus scales primarily with $T \times c_f$ when sufficient parallel resources are available. 
}

A key advantage of MPPI is that it finds the global optimal solution (in the sense of KL divergence minimization) through a single round of sampling and weighted averaging\footnote{However, as discussed in Section~\ref{sec:sample_complexity}, the required number of samples depends on the configuration.}.
This stems from the concavity of Eq.~(\ref{eq:convex_objective}), which results from fixing the covariance of the multivariate Gaussian variational distribution.
If we simultaneously optimize the covariance, Eq.~(\ref{eq:mppi_objective}) would no longer be concave, requiring iterative repetition of the three steps above in a single optimization, increasing computational cost.
However, since the variational distribution's covariance setting significantly impacts performance, careful configuration is necessary, and discussed in Section~\ref{sec:survey_prior_distributions}.

\subsection{Summary of the Tutorial}
\label{sec:tutorial_summary}
In this tutorial, we introduced the basic theory behind PI-MPC.
We conclude by summarizing the framework with the following relation:
\begin{align}
\underset{\text{Variational dist.}}{\pi^{*}_{\theta}(\myvec{u}_{0:T-1})}
 \mathrel{\overset{\text{min.\ KL div.}}{\longleftrightarrow}}
\underset{\text{Optimal control dist.}}{\pi^*(\myvec{u}_{0:T-1})} \propto \underset{\text{Boltzmann dist.}}{\exp\left(-\lambda^{-1} J_{\tau}(\myvec{u}_{0:T-1})\right)} \; \times \; \underset{\text{Prior dist.}}{p(\myvec{u}_{0:T-1})}. \label{eq:summary}
\end{align}
This relation highlights two key points. First, the optimal control distribution can be written as the product of a Boltzmann term and a prior (Eq.~(\ref{eq:optimal_control_distribution})). Second, we approximate this target distribution with a parameterized variational distribution by minimizing the KL divergence.
Control inputs are then obtained by sampling from the resulting variational distribution as in Eq.~(\ref{eq:variational_inference}).

For a given specific task and \revised{robot}{control target}, the trajectory cost $J_{\tau}$ and the control input $\myvec{u}_{0:T-1}$ are uniquely determined.
Additionally, the optimal control distribution is theoretically given by Eq.~(\ref{eq:optimal_control_distribution}). 
Therefore, the main design choices are the prior $p(\myvec{u}_{0:T-1})$, and the variational family $\pi_{\theta}$.
Based on this framework, we organize the challenges and methods in PI-MPC as follows.

\section{Survey of Probabilistic Inference-Based MPC Methods}
\label{sec:survey}

As established in Section~\ref{sec:tutorial_summary}, the main design choices in PI-MPC reduce to two elements: the prior distribution $p(\mathbf{u}_{0:T-1})$, and the variational family $\pi_{\boldsymbol{\theta}}$.
Each choice involves distinct trade-offs between solution quality, computational cost, and robustness.
This section surveys the key challenges that arise when instantiating these choices for real-world \revised{robotic tasks}{systems}, together with representative methods that address them.

We organize the discussion as follows.
Sections~\ref{sec:survey_prior_distributions}--\ref{sec:survey_constraints} address algorithmic design challenges: prior distribution design (Section~\ref{sec:survey_prior_distributions}), multi-modal distribution handling (Section~\ref{sec:survey_multimodal}), and constraint satisfaction (Section~\ref{sec:survey_constraints}).
Sections~\ref{sec:survey_high_dof}--\ref{sec:survey_hardware} turn to practical scalability, covering high Degree-of-Freedom (DoF) robots and hardware acceleration.
Finally, Sections~\ref{sec:survey_theory}--\ref{sec:survey_connections} examine theoretical foundations and connections to related paradigms.

\subsection{Design of Prior Distributions}
\label{sec:survey_prior_distributions}

The prior distribution $p(\mathbf{u}_{0:T-1})$ plays two conceptually equivalent roles in PI-MPC: it defines the sampling region from which candidate solutions are drawn, and it acts as a regularizer that shapes the notion of optimality through the KL penalty in Eq.~(\ref{eq:kl_divergence_min}).
A poorly chosen prior wastes samples in low-probability regions of the optimal distribution, while an overly narrow prior risks missing the global optimum entirely.
The principal lever for addressing this trade-off is to bring the prior close to the modes of the Boltzmann distribution while keeping its variance small enough for efficient sampling yet large enough to avoid premature convergence.

Existing methods can be grouped into four categories: (i) shaping the prior for a better exploration--exploitation balance, (ii) guiding samples with auxiliary controllers, (iii) adapting the prior online in a cost-aware manner, and (iv) learning the prior from data.

\subsubsection{What is a Good Prior Distribution?}
\begin{figure}[t]
    \centering
    \includegraphics[width=0.95\linewidth]{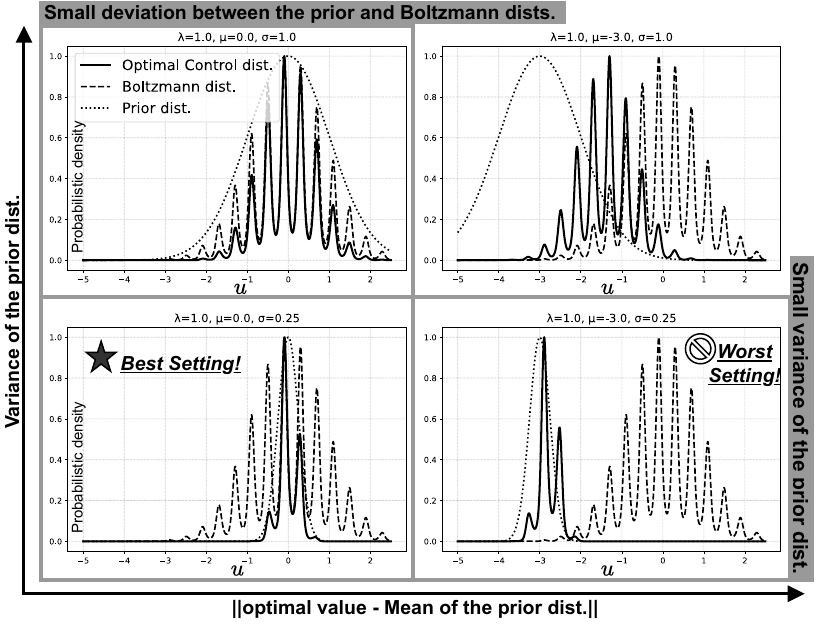}
    \caption{Effect of prior distribution mismatch and variance on the optimal control distribution.
            Larger deviation between the prior and the Boltzmann distribution deteriorates optimality, while increased prior variance mitigates this effect when the mismatch is large.
            Priors located near the Boltzmann mode with small variance yield the most favorable performance, whereas distant low-variance priors produce the poorest outcome.
            The cost function is the same as in Fig.~\ref{fig:cost_and_boltzman}, and four Gaussian prior distributions $p(u)=\mathcal{N}(\mu, \sigma), \mu \in \{-3.0, 0.0\}$ and $\sigma \in \{0.25, 1.0\}$ are compared.
            }
    \label{fig:prior_tuning}
\end{figure}

Before reviewing specific methods, we establish the design objective using a concrete example.
Figure~\ref{fig:prior_tuning} illustrates how the optimal control distribution varies with the mean and variance of the prior.
When the prior mean is far from the principal mode of the Boltzmann distribution, the resulting optimal control distribution shifts away from low-cost regions, degrading optimality (bottom-right and top-right panels).
Increasing the prior variance partially compensates for this mismatch by spreading samples more broadly (top-right), but when the prior is already well-centered, a large variance dilutes probability mass around the mode and reduces the chance of generating near-optimal actions (top-left).
The worst case is a prior that is both far from the Boltzmann mode and tightly concentrated (bottom-right); the best case is a prior centered on the mode with small variance (bottom-left).

From an algorithmic standpoint, this picture is intuitive: in the MPPI algorithm (Section~\ref{sec:mppi_derivation}), the prior defines the \emph{search region} from which all samples are drawn.
Focusing this region around the modes of the Boltzmann distribution maximizes both sample efficiency and solution quality, which is supported by theoretical analyses~\citep{Yoon2022-jb,Tao2022-dj}.
The methods reviewed below all aim, in different ways, to achieve this ideal: a prior with small variance positioned near an optimal mode (Fig.~\ref{fig:prior_tuning}, bottom-left).

\subsubsection{Balancing Exploration and Exploitation via Prior Distribution}

Reducing the variance of a standard Gaussian prior improves sample efficiency and solution quality, but it also narrows the search region.
If the prior is misaligned with the true mode, a small variance can trap in a suboptimal region and slow convergence.
The key idea is therefore to reshape the prior so that its variance is small in a meaningful sense, \ie, concentrating samples around promising regions, while retaining sufficiently heavy tails to avoid missing distant modes.

One family of methods achieves this by replacing the Gaussian prior with a leptokurtic distribution, \ie, one with higher kurtosis and heavier tails.
\citet{mohamed2022autonomous} used a normal--log-normal (NLN) mixture, and \citet{vlahov2024low} employed colored noise as the prior.
Such distributions suppress high-frequency components while preserving low-frequency structure, yielding temporally smooth samples that concentrate around plausible trajectories without sacrificing coverage.

Temporal smoothness can also be imposed more directly.
Post-processing filters~\citep{andrejev2025pi} and spline interpolation~\citep{bhardwaj2022storm} smooth samples after generation, while other methods redefine the action space in terms of action derivatives~\citep{kim2022smooth,lee2time}, so that sampling in the new space inherently produces smooth trajectories.
A more principled alternative is to embed temporal correlations into the inference problem itself: Gaussian Processes (GPs) naturally encode such correlations and can replace the temporally independent Gaussian prior used in standard MPPI~\citep{mukadam2018continuous,urain2022learning,watson2023inferring}.

\subsubsection{Guiding Samples with Auxiliary Controllers}
Rather than reshaping the prior itself, another strategy is to steer samples toward promising regions using feedback from auxiliary controllers.
Each sampled trajectory is corrected by a lightweight controller that pushes it toward a mode of the optimal control distribution, effectively narrowing the sampling region without reducing the nominal prior variance.

When linear approximations of dynamics and cost are available, analytic feedback controllers offer a computationally cheap solution.
Examples include task-specific Linear Quadratic Regulators~\citep{trevisan2024biased}, iterative Linear Quadratic Gaussian (iLQG) controllers~\citep{williams2018robust,gandhi2021robust}, and covariance steering controllers~\citep{yin2022trajectory}.
For problems where such approximations are unavailable, gradient-based methods provide a more general alternative: convex optimization~\citep{rastgar2024priest}, Stochastic Gradient Descent~\citep{bharadhwaj2020model}, and Adam~\citep{heetmeyer2023rpgd} can guide samples using only gradient information of the cost and dynamics.
For the most general setting, \ie, where even cost and dynamics gradients may not be available, Stein Variational Gradient Descent (SVGD) offers a derivative-free option that approximates the KL gradient from samples alone~\citep{lambert2021stein,honda2024stein,miura2024spline}.

\subsubsection{Cost-Aware Adaptive Prior Distribution}
Instead of correcting individual samples, a complementary strategy is to adapt the prior distribution itself in response to observed trajectory costs.
Because the adaptation modifies the distribution rather than individual samples, its computational cost is independent of the sample count, and the resulting algorithm remains consistent with the PI-MPC theoretical framework.

The central difficulty is a circular dependency: evaluating the quality of a prior requires drawing i.i.d.\ samples and rolling them out to compute trajectory costs, but these costs are themselves influenced by the prior.
\citet{asmar2023model} break this circularity through iterative refinement, \ie, alternating between cost evaluation and prior update, at the expense of higher computational cost.
Lighter-weight alternatives avoid full iteration: \citet{tao2022path} uses Stochastic Control Barrier Functions (SCBF) to set a safe mean and variance, while \citet{honda2024stein} employs SVGD on a small set of pilot samples to capture the mode shape before the main sampling phase.
Other approaches sidestep the circularity by predicting future cost characteristics: \citet{mohamed2025towards} propagate variance dynamics via the unscented transform and fold the result into the cost, and \citet{yi2024covo} derive a closed-form optimal covariance under simplifying assumptions on the cost structure.

\subsubsection{Learning-based Prior Distribution}
\revised{When offline data or simulation}{When offline or simulation data} is available, machine learning can be used to construct priors that are already well-aligned with the task before online execution begins.
If expert demonstrations are available, generative models trained on the dataset can serve directly as priors.
\citet{urain2022learning} learn an energy-based model of the optimal trajectory distribution, while \citet{carvalho2023motion} uses diffusion models~\citep{ho2020denoising} to capture richer multi-modal structure.
A complementary line of work incorporates model-free RL policies as priors: pre-trained policies can initialize the MPPI solution~\citep{qu2023rl,wang2025residualmppi}, or MPPI can be embedded as a residual correction during RL training so that the learned policy and the MPC prior co-evolve~\citep{sacks2024deep}.
In model-based RL, MPPI serves as the planner while a model-free policy provides the prior~\citep{hansen2022temporal}.

More specialized approaches learn environment-specific priors using Normalizing Flows~\citep{power2022variational,sacks2023learning}.
By incorporating the KL objective in Eq.~(\ref{eq:kl_divergence_min_wip}) into the training loss, these methods can learn priors in a self-supervised manner without expert data, conditioning only on environment features and start--goal pairs.

\subsection{Handling of Multi-Modal Distributions}
\label{sec:survey_multimodal}

Multi-modality of the optimal control distribution arises naturally in \added{physical systems, such as}
robotics, whenever multiple solutions have comparable costs: an autonomous vehicle can pass an obstacle on either side and a manipulator can choose among several grasp poses.
Failing to handle multi-modality can produce dangerous \emph{mode-averaging} behavior. For example, steering directly into an obstacle rather than clearly committing to one side.
The principal lever is the choice of variational family $\pi_{\boldsymbol{\theta}}$ and the divergence used to fit it.
Existing approaches fall into two categories: (i) approximating the full multi-modal distribution with an expressive variational family, and (ii) deliberately seeking a single mode for decisive action selection.
The former is preferable when exploration and diversity are valuable (\eg, during RL), while the latter suits deployment scenarios that demand immediate, unambiguous decisions.

\subsubsection{Variational Inference with Multi-modal Variational Distributions}
\label{sec:multimodal_vi_dist}
The most direct strategy is to replace the unimodal Gaussian of MPPI with a more expressive variational family.
\citet{osa2020multimodal} formulated offline trajectory optimization as cost-weighted density estimation using Gaussian Mixture Models (GMMs), and \citet{okada2020variational} adopted GMMs as the variational distribution within the PI-MPC framework, optimizing via mirror gradient descent.
An alternative to parametric mixtures is to represent the variational distribution nonparametrically as a set of particles updated via SVGD~\citep{lambert2021stein,barcelos2021dual}.
\citet{wang2021variational} generalized these particle-based methods by replacing the KL divergence with the Tsallis divergence, providing additional flexibility in controlling mode-covering behavior.
Outside the strict VI framework, \citet{carvalho2023motion} integrated diffusion models into PI-MPC to generate multi-modal trajectory samples.

The primary strength of these methods is their ability to maintain solution diversity, which enhances exploration and can improve learning efficiency in model-based RL~\citep{okada2020variational}.
Their main weakness relative to MPPI is computational: as shown in Eq.~(\ref{eq:mppi_mu}), MPPI obtains the globally optimal Gaussian approximation through a single round of sampling and weighted averaging, whereas multi-modal VI methods require iterative parameter updates, increasing both computational cost and the risk of numerical instability\footnote{MPPI also requires iterative updates when simultaneously optimizing the mean and covariance~\citep{wang2021variational}.}.

\subsubsection{Seeking a Single Mode in Multi-modal Distributions}

When the goal is to execute a single action sequence, approximating the entire multi-modal distribution may be unnecessary.
Instead, one can target a single high-quality mode and commit to it.

Clustering-based methods achieve this by partitioning samples post hoc.
\citet{park2025csc} employed DBSCAN to cluster samples by density, computed a weighted average within each cluster, and selected the cluster representative with the lowest cost.
\citet{zhang2024multi} computed separate weighted averages for multiple cost components and then re-weighted all samples to fuse them into a single control sequence.

A more principled alternative is to change the divergence objective.
Standard PI-MPC minimizes the \emph{forward} KL divergence $D_{\mathrm{KL}}(\pi^* \| \pi_{\boldsymbol{\theta}})$, which encourages the variational distribution to cover all modes (mode-covering).
Minimizing the \emph{reverse} KL divergence $D_{\mathrm{KL}}(\pi_{\boldsymbol{\theta}} \| \pi^*)$ instead yields a mode-seeking solution that concentrates on a single mode.
Because the reverse KL does not admit a closed-form solution, iterative methods such as SVGD~\citep{honda2024stein} and mirror descent~\citep{kobayashi2022real} are used in practice.

As discussed in Section~\ref{sec:symmetry_breaking}, symmetry breaking causes the number of modes to decrease over time, so even mode-covering methods like MPPI eventually converge to a single mode.
However, in dynamic environments or time-critical scenarios, this natural convergence may be too slow, and explicit mode-seeking can prevent dangerous hesitation.

\subsection{Handling of Constraints}
\label{sec:survey_constraints}

Real-world optimal control problems invariably involve constraints on both states and control inputs.
Control input constraints (\eg, actuator limits) are relatively easy to enforce: samples that exceed bounds can be clipped after drawing from the prior~\citep{williams2018information}, or inputs can be mapped through saturating functions such as the hyperbolic tangent function; either approach can be absorbed into the dynamics of Eq.~(\ref{eq:dynamics}).
State constraints (\eg, collision avoidance, joint limits) are fundamentally harder because they couple the constraint with the forward dynamics and cannot be enforced by modifying a single sample in isolation.
The simplest remedy is to add penalty terms to the cost function in Eq.~(\ref{eq:cost_func})---PI-MPC readily accommodates discontinuous penalties such as indicator functions and cost maps~\citep{macenski2020marathon2}---but penalties cannot guarantee strict satisfaction: if all samples from the prior violate a constraint, the ``optimal'' solution will itself be infeasible.
Methods for stricter enforcement fall into two categories: (i) projecting samples onto the feasible set during prior sampling, and (ii) incorporating constraints directly into the variational inference problem.
Projection methods are simpler to implement but may require differentiability or linearity assumptions; constrained VI is more principled but computationally heavier.

\subsubsection{State-constrained Prior Sampling}
The idea is to modify samples \emph{before} cost evaluation so that they satisfy state constraints.
In the MPPI framework, this typically means drawing Gaussian samples, rolling out the dynamics, and then projecting any constraint-violating trajectories onto the feasible set.

Several projection mechanisms have been proposed.
Control Barrier Functions~\citep{gandhi2023safe,gandhi2022safety} project inputs to enforce forward-invariance of a safe set; primal--dual interior-point methods~\citep{park2025csc} and convex optimization~\citep{rastgar2024priest} solve small local programs to find the nearest feasible input.
\citet{yan2024output} take the inverse approach: they sample directly in the state space and convert feasible state trajectories to control inputs via inverse dynamics.

For settings that require future safety guarantees, \citet{borquez2025dualguard} project samples onto a future safety region defined by Hamilton--Jacobi reachability analysis, ensuring that the system can always return to a safe state.
\citet{wang2025constrained} extends the framework to hierarchical equality and inequality constraints by combining the Lagrange multiplier method with null-space projection.

Projection-based methods are straightforward to implement and can enforce constraints strictly when the projection itself is exact.
However, they often require differentiability or linearity of the constraint functions and dynamics, which limits the generality of the PI-MPC formulation.
Moreover, projection distorts the sample density, potentially biasing the importance weights; the resulting solution may therefore deviate from the true constrained optimum obtained by the VI-based approach described next.

\subsubsection{Constrained Particle-based Variational Inference}

A more principled alternative is to embed state constraints directly into the variational inference problem, yielding a \emph{constrained} KL minimization over Eq.~(\ref{eq:variational_inference}).
This line of work builds on constrained VI methods from the machine learning literature~\citep{liu2021sampling,zhang2022sampling}, often implemented via SVGD~\citep{liu2016stein}, which represents the target distribution as particles and updates them along the gradient of the KL divergence.

\citet{power2024constrained} extended the constrained SVGD framework of \citet{zhang2022sampling} to handle multiple linear and nonlinear constraints, applying it to robotic planning problems by projecting particle updates onto the tangent space of the constraint manifold.
\citet{tabor2025constrained} proposed a more unified formulation that folds constraints into the SVGD kernel, avoiding explicit tangent-space projection.

Constrained VI is theoretically appealing: it directly approximates the optimal control distribution conditioned on constraint satisfaction and can, in principle, handle constraints on distributional statistics (\eg, expected constraint satisfaction).
In practice, however, iterative particle updates incur significant computational overhead, and the additional hyperparameters (\eg, step size, constraint penalty weights) make tuning more delicate than for standard MPPI.

\subsection{Applications to High-Dimensional Robots}
\label{sec:survey_high_dof}
Early PI-MPC studies focused on low-dimensional systems such as ground vehicles with simple action spaces.
As the field moves toward more capable platforms, \eg, humanoids with 30+ joints, multi-fingered hands, and soft robots, the dimensionality of the control problem grows rapidly, and naive Monte Carlo sampling becomes prohibitively expensive.
Scaling PI-MPC to high degrees of DoF requires addressing two coupled challenges: the exponential growth of sample complexity with action-space dimension, and the difficulty of obtaining accurate dynamics models for systems with complex contact and deformation.
This section reviews how recent work tackles these challenges for platforms including 4-Wheel Independent Drive and Steering (4WIDS) vehicles~\citep{aoki2024switching}, manipulators~\citep{bhardwaj2022storm}, quadruped robots~\citep{keshavarz2025control}, and humanoid robots~\citep{xue2024full}.

\subsubsection{Sample Complexity in High DoF Robots}
The curse of dimensionality is the central obstacle: as the action-space dimension grows, the volume of the sampling region expands exponentially, and the fraction of samples that land near an optimal mode shrinks correspondingly.
Standard MPPI with joint-angle or joint-torque actions often fails to find good solutions within a feasible sample budget for systems.

The most common countermeasure is dimensionality reduction.
Defining the action space in terms of spline control points, rather than per-timestep joint commands, reduces the effective dimensionality and has enabled PI-MPC on various humanoid platforms~\citep{howell2022predictive,pezzato2025sampling,miura2024spline}.
Combining Halton sequences (for low-discrepancy coverage) with spline interpolation further improves sample efficiency~\citep{bhardwaj2022storm,pezzato2025sampling}, though such smoothing techniques inevitably sacrifice some optimality by restricting the space of representable trajectories.

An alternative to dimensionality reduction is iterative refinement.
Diffusion-style annealing~\citep{pan2024model} and adaptive importance sampling (AIS)~\citep{asmar2023model} sharpen the sample distribution over multiple passes, concentrating samples in high-quality regions.
These methods have shown promising results, including real-world deployment on quadruped~\citep{keshavarz2025control} and humanoid~\citep{xue2024full} robots, but the additional iterations increase computational cost.

\subsubsection{Dynamics Modeling for High DoF Robots}
Accurate dynamics models are essential for PI-MPC, yet analytical modeling becomes intractable for high-DoF robots, especially those involving contact (\eg, legged locomotion, manipulation).
A practical solution is to use GPU-accelerated physics simulators, \eg, MuJoCo~\citep{todorov2012mujoco}, Isaac Lab~\citep{mittal2025isaaclab}, and Brax~\citep{freeman1brax}, for forward rollouts during planning~\citep{howell2022predictive,pezzato2025sampling,xue2024full}.
Isaac Lab and Brax are particularly well-suited to PI-MPC because they support massively parallel simulation on GPUs, and their randomizable environments enable domain randomization for robustness.

An alternative is to learn the dynamics model from data and use it for planning, \ie, the model-based RL.
Methods such as TD-MPC~\citep{hansen2022temporal,hansentd} learn compact latent-space dynamics and plan with MPPI, achieving strong performance even on high-DoF continuous-control tasks.

\subsection{Hardware Acceleration}
\label{sec:survey_hardware}
Real-time PI-MPC demands that all computation, \ie, sampling, rollout, cost evaluation, and weight update, be completed within a few milliseconds.
Because the sampling and rollout steps are embarrassingly parallel, the primary lever for meeting this budget is hardware-level parallelism.
The choice of platform depends on the deployment context: CPUs suffice for low-dimensional systems with moderate sample counts, GPUs are the current workhorse for high-dimensional problems, and emerging architectures (\eg, FPGAs, quantum-inspired hardware) target ultra-low-latency or energy-constrained settings.

On the CPU side, the ROS~2 Navigation stack includes an MPPI-based local planner~\citep{macenski2020marathon2} implemented with the Eigen library, offering efficient single-threaded and SIMD-parallel execution suitable for embedded deployment.
On the GPU side, machine-learning frameworks such as PyTorch and JAX provide high-level abstractions that make GPU-accelerated MPPI straightforward to implement~\citep{bou2023torchrl}, though Python-level overhead limits throughput.
Native CUDA kernel implementations in C++ eliminate this overhead and deliver substantially higher performance~\citep{vlahov2024mppi}.

Beyond conventional CPUs and GPUs, recent work has begun exploring unconventional accelerators.
\citet{werthen2025ising} proposed solving the MPPI optimization on Ising machines, a form of quantum-inspired computing.
\citet{tanguy2026domain} investigated dedicated FPGA and ASIC designs tailored to the MPPI dataflow.

\subsection{Theoretical Analysis of MPPI}
Despite its empirical success, PI-MPC lacks the mature theoretical toolkit available for gradient-based MPC.
We must often tune parameters by trial and error, without formal guidance on how close the resulting solution is to the true optimum or how many samples are sufficient.
The two central questions are: (i)~how large is the \emph{optimality gap} between the MPPI solution and the solution of the original optimal control problem, and (ii)~under what conditions does MPPI \emph{converge}, and at what rate?
Progress on these questions not only informs algorithm design but is also essential for certifying PI-MPC in safety-critical applications.

\subsubsection{Optimality Gap from Original Optimal Control Problem}
Because MPPI minimizes the KL objective in Eq.~(\ref{eq:kl_divergence_min}) rather than the original cost in Eq.~(\ref{eq:ocp}), its solution can deviate from the true optimum.
\citet{williams2018information} identified a special case in which this gap vanishes: when the cost separates into state-dependent and input-dependent terms and the input cost is a quadratic weighted by the prior covariance, the MPPI solution coincides exactly with the optimal solution of the original problem.
\citet{trevisan2024biased} generalized this result, showing that for any variational family, adding an appropriate bias correction to the cost function eliminates the gap.

More broadly, the optimality gap can be traced to the prior regularization term in Eq.~(\ref{eq:kl_divergence_min}).
Appropriate prior design (Section~\ref{sec:survey_prior_distributions}) and temperature tuning can therefore reduce the gap in practice.
\citet{yi2024covo} derived a closed-form covariance that maximizes the convergence rate, directly linking prior design to optimality.
\citet{homburger2025optimality} provided a more general analysis for deterministic problems, proving that as $\lambda \to 0$ the MPPI solution converges to the deterministic optimum and bounding the control-input and value-function errors as functions of $\lambda$.

\subsubsection{Convergence Analysis}
\label{sec:survey_theory}
\citet{yi2024covo} proved that MPPI converges for quadratic costs in the infinite-sample limit, with the convergence rate depending on the temperature parameter and the prior covariance.
They further suggested that contraction properties extend to strongly convex costs with nonlinear dynamics.

Because MPPI is a zeroth-order method, its convergence is intimately tied to sample complexity.
\citet{Yoon2022-jb} and \citet{tao2022path} derived probabilistic bounds on the estimation error of the optimal distribution, showing that smaller prior covariance and lower expected trajectory costs both reduce the number of samples needed.
However, convergence depends jointly on sample number, temperature, prior covariance, and cost landscape in ways that are not yet fully disentangled.

\added{We note that the convergence and optimality results reviewed above do not directly imply closed-loop stability in the sense of classical MPC theory. Establishing stability guarantees for PI-MPC remains an open problem.}

\subsection{Connection to Other Paradigms}
\label{sec:survey_connections}

PI-MPC does not exist in isolation; it shares deep structural connections with RL, diffusion models, and data-driven learning.
Understanding these connections clarifies which tools from neighboring fields can be imported and where PI-MPC offers unique advantages.
We highlight three principal threads: the Control-as-Inference bridge to RL, the emerging link between MPPI and diffusion processes, and the use of PI-MPC as an engine for data generation and policy learning.

\paragraph{Control as Inference and Reinforcement Learning}
The Control-as-Inference framework~\citep{levine2018reinforcement} provides a unified probabilistic perspective that encompasses both PI-MPC and RL.
Within this framework, RL algorithms such as Soft Actor-Critic (SAC)~\citep{haarnoja2018soft} and Maximum a Posteriori Policy Optimization (MPO)~\citep{abdolmaleki2018maximum} emerge from the same KL-regularized inference that underlies PI-MPC; the key difference is that RL methods amortize the optimization over a learned policy, whereas PI-MPC solves it online at each time step.
Model-based RL bridges the two: methods such as PETS~\citep{chua2018deep}, Dreamer~\citep{hafner2019learning}, and TD-MPC~\citep{hansen2022temporal,hansentd} learn dynamics and reward models while using PI-MPC planners (CEM or MPPI) for action selection.
In this context, the stochastic nature of PI-MPC naturally provides the exploration variability that model-based RL requires.
Empirically, \citet{wangbootstrapped} report that, on more challenging tasks, incorporating MPPI into a model-based RL pipeline leads to more robust value-function learning and improved control performance over a model-free SAC-style policy alone.

\paragraph{Diffusion Models}
MPPI and score-based diffusion processes~\citep{ho2020denoising} are closely related: both iteratively refine a noisy distribution toward a target.
\citet{pan2024model} and \citet{xue2024full} showed that when an optimal control problem replaces training data as the target, MPPI computations can substitute for portions of the score function, enabling sample generation at each diffusion step without training.
This diffusion-style annealing captures finer modal structure than single-pass MPPI and has demonstrated strong results on high-dimensional locomotion tasks.
\citet{li2025unifying} further attempts to unify MPPI, RL, and diffusion models under the lens of gradient ascent on smoothed energy with respect to Gibbs measures, offering a common theoretical foundation.

\paragraph{Data Generation}
The stochastic nature of PI-MPC also makes it a natural fit for data augmentation and sample-efficient learning. In visual navigation, \citet{inglin2026less} uses MPPI rollouts to augment training data for imitation learning-based navigation. PI-MPC can further be coupled with learned dynamics models to generate large amounts of synthetic experience: \citet{morgan2021model} run MPPI on a learned world model~\citep{ha2018world} to produce model-generated trajectories that are mixed with real-environment data for off-policy RL, while \citet{wangcoplanner} design a dual strategy in which the world model is trained by sampling transitions that reduce epistemic uncertainty, whereas the MPPI cost function is modified at deployment to actively steer the agent into uncertain regions for more informative data collection. While the above approaches use MPPI to produce training data, it can also guide policy learning directly: \citet{wangbootstrapped} distill the Gaussian action distribution of an MPPI planner into a model-free policy, achieving high training efficiency by bootstrapping model-free RL with model-based learning and control.

\section{Conclusion}
This paper has presented a unified tutorial and survey on probabilistic inference-based Model Predictive Control (PI-MPC).
In the tutorial, we showed that the optimal control distribution admits a simple and interpretable form, \ie, the product of a Boltzmann distribution and a prior, and that practical controllers such as MPPI can be derived as variational approximations to this distribution.
In the survey, we organized the growing body of PI-MPC research around the key design choices that practitioners face: how to shape the prior distribution for efficient sampling, how to handle multimodal and constrained optimal control distributions, how to scale to high-dimensional robotic systems and real-time deployment, and how PI-MPC connects to related theoretical and learning-based paradigms.

\paragraph{Future directions}
Given the recent success of learning-based control, \eg, large-scale reinforcement learning, PI-MPC can offer additional value as a complementary component rather than a competing alternative.
Because PI-MPC can be connected to reinforcement learning and diffusion-based generative processes through a unified variational inference framework, as discussed in Section~\ref{sec:survey_connections}, it may provide a useful lens for analyzing these methods and for developing clearer theoretical interpretations and guarantees.
Moreover, purely learning-based controllers often require extensive training and can be less flexible in incorporating safety constraints or structured environment knowledge.
Hybrid approaches that integrate PI-MPC with learning-based methods therefore have the potential to combine the strengths of both paradigms: fast online adaptation and constraint-aware planning from PI-MPC, together with representation learning and generalization from data-driven models.

\bibliography{references}
\bibliographystyle{unsrtnat}

\end{document}